\documentclass{article}

\usepackage[preprint]{neurips_2026}

\usepackage[utf8]{inputenc} % allow utf-8 input
\usepackage[T1]{fontenc}    % use 8-bit T1 fonts
\usepackage{hyperref}       % hyperlinks
\usepackage{url}            % simple URL typesetting
\usepackage{booktabs}       % professional-quality tables
\usepackage{amsfonts}       % blackboard math symbols
\usepackage{nicefrac}       % compact symbols for 1/2, etc.
\usepackage{microtype}      % microtypography
\usepackage{xcolor}         % colors
\usepackage{graphicx}
\usepackage{amsmath}
\usepackage{colortbl}
\usepackage{soul}
\usepackage{color}
\usepackage{wrapfig}
\usepackage{threeparttable}
\usepackage{caption}
\usepackage{bm}
\usepackage{multirow}
\usepackage{float}

\newcommand{\modelname}{EponaV2}

\title{\modelname: Driving World Model with Comprehensive Future Reasoning}

\author{%
Jiawei Xu\footnotemark[1]\ ~\footnotemark[2]\ 
, \ 
Zhizhou Zhong\footnotemark[2]\ ~\footnotemark[3]\ , \ 
Zhijian Shu\footnotemark[2]\ ~\footnotemark[4]\ , \ 
Mingkai Jia\footnotemark[2]\ ~\footnotemark[3]\ , \ 
Mingxiao Li\footnotemark[2]\ , \\
\textbf{Jia-Wang Bian}\footnotemark[5]\ \textbf{,} \
\textbf{Qian Zhang}\footnotemark[2]\ \textbf{,} \ 
\textbf{Kaicheng Zhang}\footnotemark[6]\ \textbf{,} \ 
\textbf{Jin Xie}\footnotemark[7]\ 
\textbf{,} \
\textbf{Jian Yang}\footnotemark[1]\ ~\footnotemark[7]\ 
\textbf{,} \ 
\textbf{Wei Yin}\footnotemark[2]\ 
\\
\footnotemark[1] \ PCA Lab, VCIP, College of Computer Science, Nankai University
\footnotemark[2] \ Horizon Robotics
\footnotemark[3] \ HKUST\\
\footnotemark[4] \ NJUPT
\footnotemark[5] \ NTU
\footnotemark[6] \ Anyverse
\footnotemark[7] \ School of Intelligence Science and Technology, Nanjing University}
\begin{document}

% \renewcommand{\thefootnote}{\fnsymbol{footnote}}
% \footnotetext[1]{Corresponding authors.}

\maketitle

\begin{abstract}

% problem of perception-based models -> problem of perception-free models -> our contributions
Data scaling plays a pivotal role in the pursuit of general intelligence.
However, the prevailing perception-planning paradigm in autonomous driving relies heavily on expensive manual annotations to supervise trajectory planning, which severely limits its scalability.
Conversely, although existing perception-free driving world models achieve impressive driving performance, their real-world reasoning ability for planning is solely built on next frame image forecasting.
Due to the lack of enough supervision, these models often struggle with comprehensive scene understanding, resulting in unsatisfactory trajectory planning.
In this paper, we propose \modelname, a novel paradigm of driving world models, which achieves high-quality planning with comprehensive future reasoning.
Inspired by how human drivers anticipate 3D geometry and semantics, we train our model to forecast more comprehensive future representations, which can be additionally decoded to future geometry and semantic maps.
Extracting the 3D and semantic modalities enables our model to deeply understand the surrounding environment, and the future prediction task significantly enhances the real-world reasoning capabilities of \modelname, ultimately leading to improved trajectory planning.
Moreover, inspired by the training recipe of Large Language Models~(LLMs), we introduce a flow matching group relative policy optimization mechanism to further improve planning accuracy.
The state-of-the-art~(SOTA) performances of \modelname~among perception-free models on three NAVSIM benchmarks~(\textit{+1.3PDMS, +5.5EPDMS}) demonstrate the effectiveness of our methods.
% The code will be released upon acceptance.
\url{https://github.com/JiaweiXu8/EponaV2}.

\end{abstract}

\section{Introduction}

% data-scaling importance -> problem in autonomous driving -> current perception-based models
Data scaling laws~\cite{scaling-nlp, scaling-vit, scaling-vit22} have proven highly effective in developing neural networks with general intelligence.
However, autonomous driving models, which must make trajectory planning decisions across a wide variety of scenarios, still suffer from the high costs associated with training data production.
The primary challenge in scaling training data is the expense of manual perception annotations, such as object bounding boxes and semantic segmentations, which are crucial for perception-based models utilizing the popular perception-planning paradigm.
Recently, driving models that rely on these manual labels have seen significant improvements in both Bird's-Eye-View~(BEV) methods~\cite{uniad, vad, tranfusers-v1, wote, law, diffusiondrive, goalflow, bridgedrive, meanfuser, guideflow} and Vision-Language-Model~(VLM)~\cite{drivevlm, orion, autovla, recogdrive, vggdrive, cogdriver, colavla, sgdrive} based approaches.
% Recently, driving models that rely on these manual labels have seen significant improvements in both Bird's-Eye-View~(BEV) methods~\cite{uniad, think-twice, vad, driveadapter, genad, hydra-mdp, vadv2, causalvad, tranfusers-v1, transfuser, paradrive, momad, drivedpo, wote, law, synad, hipad, colmdriver, hydra-next, seerdrive, pgs, rawdrive, e3ad, gaussianfusion, diffe2e, resworld, diffusiondrive, goalflow, bridgedrive, meanfuser, safedrive, spatial-geo, resad, knowval, guideflow, percept-wam} and Vision-Language-Model~(VLM)~\cite{drivevlm, orion, curious-vla, reflectdrive, autodrive-r2, drivemamba, real-ad, vlr-driver, futuresightdrive, simlingo, drivelm, dima, autodrive-p3, autovla, recogdrive, vggdrive, cogdriver, colavla, sgdrive, spacedrive, minddriver, linkvla, elf-vla, drivemoe, wam-flow, lcdrive} based approaches.
Nevertheless, these methods typically depend on manual annotations to design auxiliary tasks, such as Occupancy Prediction~(Occ-Pred) and Visual Question Answering~(VQA), to build their perception capabilities.
This heavy reliance on expensive manual labeling makes it difficult to apply data scaling laws to perception-based driving models, limiting their potential for stronger planning abilities.

\begin{figure}[!t]
    \centering
    \includegraphics[width=\linewidth, trim=0 421 570 0, clip]{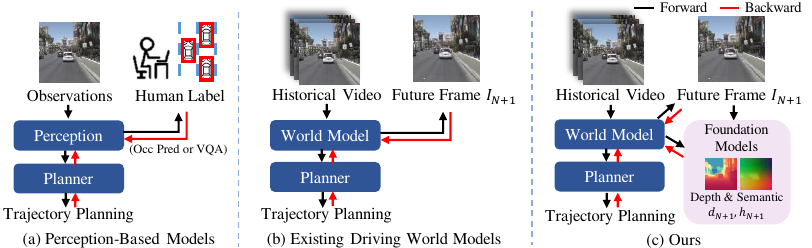}
    \caption{Training Pipeline Comparison. (a)~Perception-based models require manual labels to build perception abilities for planning. (b)~Existing perception-free driving world models employ limited future reasoning for planning only by future frame prediction. (c)~Ours utilizes foundation models for comprehensive future reasoning by future depth and semantic predictions for better planning.}
    \label{fig:comparison}
\end{figure}

% perception-free models -> existing problems (limit supervision)
Consequently, perception-free models have been introduced to eliminate the need for manual perception labels during training~\cite{drivinggpt, robotron-sim, world4drive, epona, drivevla-w0, pwm, drivelaw}.
However, lacking manual perception supervision, these models often struggle to fully comprehend complex real-world environments.
To address this, training world models to reason the representations of the future from current observations for trajectory planning has proven effective for trajectory planning~\cite{epona, drivevla-w0, pwm, drivelaw}.
Naturally, future representations that encode richer and more actionable information require a stronger understanding and reasoning ability of the world and are crucial for better planning.
As illustrated in Figure~\ref{fig:comparison}, a prevailing method to facilitate future reasoning is to train models to forecast the next video frame~\cite{epona, drivevla-w0, pwm, drivelaw}.
This raises a critical question: \textbf{\textit{Is the future reasoning ability built solely via next-frame image prediction sufficient for trajectory planning?}}
We argue that it is insufficient.
The information critical for planning within raw images is highly entangled and difficult to utilize directly.
Therefore, inferred representations supervised purely by image prediction inherit these ambiguities, posing challenges for better future reasoning ability and trajectory planning.

% Unfortunately, these future frames often lack sufficient explicit 3D and semantic information, both of which are critical for driving decision making, resulting in an inadequate understanding of the surroundings.
% Certain methods~\cite{world4drive} incorporate depth-aware inputs for 3D aggregation and leverage self-supervised present-frame segmentation for semantic cues.
% However, these methods probably cannot force the model to enforce reasoning over these data for specific tasks, and there is no guaranty that the model fully utilizes and understands these embedded 3D and semantic information, leading to suboptimal planning performance.
% An simple but effective way for driving models to correct the planning results is directly selecting predefined trajectory from a vocabulary~\cite{world4drive, drivevla-w0, pwm, drivelaw}.
% However, predefined trajectories not only limit the diversity of the model’s decisions but also require careful design to handle a variety of situations, which prevents this method from being widely applied.

\begin{figure}[t]
    \centering
    \includegraphics[width=\linewidth, trim=0 410 490 0, clip]{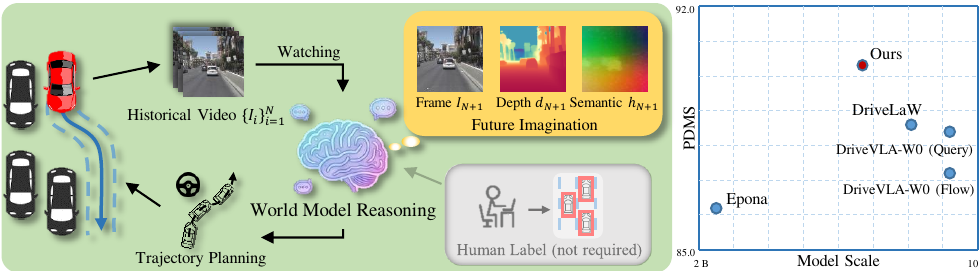}
    \caption{\modelname. Without relying on manual perception labels, our model develops a strong future reasoning ability of real-world environments for trajectory planning via comprehensive future predictions, achieving SOTA planning performance among perception-free driving world models.}
    \label{fig:teaser}
\end{figure}

% our contributions
% human inspired 3D and semantic -> utilize them in future reasoning
To address these limitations, we propose \modelname, a novel paradigm of perception-free driving world models.
Inspired by human driving behavior, where decisions rely on anticipating the distances to forward obstacles and understanding their semantic context, we posit that future reasoning for trajectory planning should also utilize both the 3D geometry and semantic information of the environment.
To this end, we introduce a simple yet effective perception-free mechanism: the prediction of future depth and semantic maps.
By enforcing future depth prediction, \modelname~comprehensively captures the 3D geometry and motion dynamics of surrounding objects.
Furthermore, forecasting future feature maps derived from large-scale segmentation models~\cite{sam3} imparts a profound semantic understanding of the scene, also ensuring that \modelname~remains focused on elements critical to driving decisions.
Without relying on manual perception labels, the comprehensive forecasting approach builds a strong real-world understanding and future reasoning ability for \modelname.
% This forecasting approach fully leverages the strong real world future reasoning and representation capabilities of the underlying VLM~\cite{qwen3-vl} and foundational DINO model~\cite{dinov3, dinotok} in our driving world model.
Consequently, the inferred future representations yield rich and actionable information for trajectory planning, substantially boosting overall driving performance.
Moreover, inspired by the training recipe of LLMs, we introduce a flow matching Group Relative Policy Optimization~(GRPO) mechanism for our flow-based planner to further improve the accuracy of the generated trajectories.
This flow matching GRPO autonomously refines the model's planning results using streamlined reward functions.
Experiments on three NAVSIM~\cite{navsim-v1, navsim-v2} benchmarks demonstrate that \modelname~achieves SOTA performance among perception-free driving models, confirming the effectiveness of our approach.

In general, the main contributions of this paper can be summarized as follows.
\begin{itemize}
    \item We propose \modelname, a novel perception-free paradigm of driving world models, which achieves high-quality planning and is much easier to perform data scaling training.
    \item We introduce a novel approach that requires \modelname~to predict future depth and semantic maps, enabling a stronger future reasoning ability for improved planning.
    \item We design the flow matching group related policy optimization mechanism to improve the accuracy of the trajectories from our flow matching planner with simple reward functions.
\end{itemize}

\section{Related Work}

% BEV -> BEV models -> VLA models
\noindent\textbf{Perception-Based Driving Models.}
Perception labels are fundamental to modern perception-planning autonomous driving paradigm, as they enable models to detect and segment surrounding objects for more effective planning.
BEVFormer~\cite{bevformer} designs the BEV representation for autonomous driving tasks~\cite{uniad, vad, vadv2, causalvad, tranfusers-v1, transfuser, momad, wote, law, synad, hipad, colmdriver, hydra-next, seerdrive, pgs, rawdrive, e3ad, gaussianfusion, diffe2e, resworld, spatial-geo, resad, guideflow, percept-wam}, which aggregates spatial positions and semantic information from the driving environment.
% BEVFormer~\cite{bevformer} designs the BEV representation for autonomous driving tasks~\cite{uniad, think-twice, vad, driveadapter, genad, hydra-mdp, vadv2, causalvad, tranfusers-v1, transfuser, paradrive, momad, drivedpo, wote, law, synad, hipad, colmdriver, hydra-next, seerdrive, pgs, rawdrive, e3ad, gaussianfusion, diffe2e, resworld, spatial-geo, resad, knowval, guideflow, percept-wam}, which aggregates spatial positions and semantic information from the driving environment.
Building on this, DiffusionDrive~\cite{diffusiondrive, diffusiondrivev2} demonstrates that diffusion models~\cite{ddpm, ddim} serve as powerful policy makers for BEV-based planning.
GoalFlow~\cite{goalflow} teaches models to predict the goal points for more accurate trajectory generation.
BridgeDrive~\cite{bridgedrive} proposes an anchor-guided diffusion bridge policy for closed-loop planning.
MeanFuser~\cite{meanfuser} introduces Gaussian mixture noise to guide trajectory generation for anchor-free planning.
SafeDrive~\cite{safedrive} performs explicit and interpretable safety reasoning to enable safer and more accurate trajectory generation.
For computational efficiency, SparseDrive~\cite{sparsedrive, bridgead, flowad, distilldrive} designs a BEV-free paradigm, while BridgeAD~\cite{bridgead} enables the effective use of historical prediction and planning to improve the coherence and accuracy of the models.
Research has also explored environmental understanding and scalability; DriveDreamer~\cite{drivedreamer} focuses on real-world comprehension, DrivoR~\cite{drivor} utilizes pretrained Vision Transformers~(ViT) with sensor-aware register tokens, and both RAP~\cite{rap} and SimScale~\cite{simscale} propose scalable data augmentation pipelines.
Recently, the success of VLMs has proven their abilities for different tasks~\cite{qwen3-vl}, and fine-tuning VLMs to Vision-Language-Action~(VLA) models for autonomous driving tasks might be a possible way~\cite{drivevlm, orion, curious-vla, reflectdrive, autodrive-r2, drivemamba, real-ad, vlr-driver, futuresightdrive, drivelm, dima, cogdriver, colavla, sgdrive, spacedrive, minddriver, linkvla, elf-vla, drivemoe, wam-flow, lcdrive}.
% Recently, the success of VLMs has proven their abilities for different tasks~\cite{qwen3-vl}, and fine-tuning VLMs to Vision-Language-Action~(VLA) models for autonomous driving tasks might be a possible way~\cite{drivevlm, orion, curious-vla, reflectdrive, autodrive-r2, drivemamba, real-ad, vlr-driver, futuresightdrive, simlingo, drivelm, dima, cogdriver, colavla, sgdrive, spacedrive, minddriver, linkvla, elf-vla, drivemoe, wam-flow, lcdrive}.
Notable examples include AutoVLA~\cite{autovla}, ReCogDrive~\cite{recogdrive}, and AutoDrive-P$^3$~\cite{autodrive-p3}, which perceive environments via a question-and-answer format and improve planning through DPPO~\cite{dppo}.
Furthermore, VGGDrive~\cite{vggdrive} integrates 3D-aware models to enhance the geometric perception of VLA frameworks.
Despite these advancements, perception-based models rely heavily on labor-intensive annotations, which hinders their ability to scale with massive datasets.

% existing models -> bad performance
\noindent\textbf{Perception-Free Driving Models.}
Unlike perception-based approaches, perception-free models eliminate the need for costly manual labels such as object bounding boxes or semantic segmentation.
DrivingGPT~\cite{drivinggpt} employs VLMs to unify simulation and planning tasks, while RoboTron-Sim~\cite{robotron-sim} enhances safety using scenario-aware prompts and an image-to-ego encoder.
To facilitate planning without explicit perception, World4Drive~\cite{world4drive} constructs a latent world model, and Epona~\cite{epona} utilizes a diffusion world model for autoregressive video generation.
Additionally, DriveVLA-W0~\cite{drivevla-w0}, DriveLaW~\cite{drivelaw} and PWM~\cite{pwm} implement dense self-supervision for planning by training on future image predictions.
While perception-free models offer advantages in data scaling, they often struggle to interpret complex environments with the limited self-supervision, which can lead to suboptimal performance.

\section{Methods}

\subsection{Applying World Model to Driving with Flow Matching}

\textbf{Input and World Model Backbone.}
Following the previous works~\cite{epona}, our \modelname~only takes $N$ frame front-view observation videos $\{I_i\}_{i=1}^N$ and the corresponding relative movements $\{\Delta A_i\}_{i=1}^N$ as input.
A strong auto-encoder can effectively extract the semantic information of an image, significantly enhancing the understanding ability of a driving world model.
Therefore, we first encode the images into feature sequences $\{F_i\}_{i=1}^N$ independently using the frozen pretrained DINO-based representation model DINO-Tok~\cite{dinov3, dinotok}.
% Subsequently, in line with recent approaches~\cite{drivevla-w0, unified-vla, interleave-vla, pwm, drivelaw}, we leverage the advanced reasoning capabilities of VLMs.
Subsequently, in line with recent approaches~\cite{drivevla-w0, unified-vla, interleave-vla, pwm, drivelaw}, we initialize the backbone of our world model with VLMs~\cite{qwen3-vl}.
By processing these flattened features alongside their associated actions, the backbone implicitly infers the corresponding future representations $\{F_i', \Delta A_i'\}_{i=1}^N$ for comprehensive world modeling:
\begin{align}
    \{F_1', \Delta A_1', F_2', \Delta A_2', ..., F_N', \Delta A_N'\} = WM(\{F_1, \Delta A_1, F_2, \Delta A_2, ..., F_N, \Delta A_N\}).
\end{align}
During processing, a causal attention mask is applied to the backbone to ensure that future observation features do not influence the current frame inference.

\textbf{Rectified Flow Matching Planner.}
Once inferred by the world model backbone, the frame-wise future representations $\{F_i', \Delta A_i'\}$ are passed to a rectified flow matching~\cite{flow-ode, flow-matching, rectified-flow, svg, svg-t2i} planner for trajectory generation.
Similar to~\cite{epona, goalflow}, the flow matching planner $\bm{v}_{traj}$ predicts the velocity from data samples $\bm{x}_0$ to Gaussian noises $\bm{x}_1\sim \mathcal{N}(\mathbf{0}, \mathbf{I})$:
\begin{align}
    d\bm{x}_t = \bm{v}_{traj}(F_i', \Delta A_i', \bm{x}_t, t)dt, t\in[0, 1].
\end{align}
During training, the process of adding noise to a data sample for the diffuse timestep $t$ is:
\begin{align}
    \bm{x}_t = (1 - t)\bm{x}_0 + t\bm{x}_1, t\in [0, 1].
\end{align}
To train the flow matching planner, the planner prediction can be supervised by the direction from the ground-truth trajectory to the noise:
\begin{align}
    L_{traj} = ||\bm{v}_{traj}(F_i', \Delta A_i', \bm{x}_t, t) - (\bm{x}_1 - \bm{x}_0)||^2.
\end{align}
Consequently, to derive the predicted trajectory $\hat{\bm{x}}_0$, the sampling process for the flow matching planner at each frame is defined as:
\begin{align}
    \hat{\bm{x}}_{t + \Delta t} = \hat{\bm{x}}_t + \bm{v}_{traj}(F_i', \Delta A_i', \hat{\bm{x}}_t, t)\Delta t, \ t\in[0, 1], \ \hat{\bm{x}}_1\sim \mathcal{N}(\mathbf{0}, \mathbf{I}).
\end{align}
Rectified flow matching planners~\cite{epona, goalflow} typically converge much faster than traditional diffusion-based alternatives~\cite{ddim, ddpm, recogdrive, diffusiondrive, diffusiondrivev2}, resulting in a more efficient and streamlined training process. 

% image prediction
\textbf{Rectified Flow Matching Future Frame Predictor.}
Decoding future representations into the next frame for supervision has proven effective in enabling world models to infer the dynamics of surrounding moving objects, thereby enhancing trajectory planning~\cite{epona, drivevla-w0, pwm, drivelaw}.
Therefore, to establish a coarse world model, following Epona~\cite{epona}, we also perform the next frame prediction of encoded images.
Specifically, we employ a rectified flow matching head $v_{img}$ conditioned on the inferred future representation $\{F_i', \Delta A_i'\}$ and a controlling relative movement $\Delta A_{i + 1}$:
\begin{align}
    L_{img} = ||v_{img}(F_i', \Delta A_i', \Delta A_{i+1}, t, F_{i+1}^t) - (\epsilon - F_{i + 1})||^2, \notag \\
    F_{i+1}^t = (1 - t)F_{i + 1} + t\epsilon,\ t \in [0, 1],\ \epsilon \sim \mathcal{N}(\mathbf{0}, \mathbf{I}).
\end{align}
During inference, the controlling relative movement $\Delta A_{i + 1}$, which is supplied as ground truth during training, is instead autonomously predicted by our trajectory planner.
% Detailed specifications regarding the architectures of our planner and future frame predictor are provided in Section~\ref{sec:suppl-impl} of the supplementary material.

\begin{figure}[t]
    \centering
    \includegraphics[width=\linewidth, trim=0 405 550 0, clip]{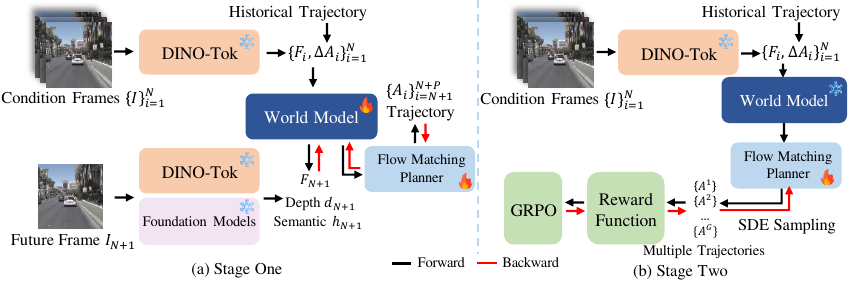}
    \caption{The pipeline of \modelname. Our model utilizes video sequences encoded by DINO-Tok~\cite{dinotok, dinov3} and corresponding historical trajectories as input. The driving world model reasons the future representations and the trajectory planner decodes these representations into trajectories. (a)~For stage one, to build the strong and comprehensive future reasoning ability of \modelname, we supervise the future image, depth and semantic maps decoded from the inferred future representations with visual foundation models. (b)~For stage two, \modelname~finetunes the predicted trajectory by flow matching GRPO~\cite{flow-grpo} with several simple reward functions.}
    \label{fig:pipeline}
\end{figure}

\subsection{Forecasting to Reason the Real World}

% advantages of perception-free models -> previous work problems -> our solution
Without the explicit guidance of manual perception labels during training, perception-free driving world models often struggle to fully comprehend complex physical environments.
Consequently, future representations inferred by these models frequently lack the rich information necessary for robust trajectory planning.
Existing methods~\cite{drivevla-w0, epona, pwm, drivelaw} typically address this limitation by only decoding the predicted future states into the next video frame for supervision.
However, we argue that relying solely on next-frame prediction is inadequate for the complexities of autonomous driving, as the visual information within raw images remains highly entangled and difficult to utilize directly.
Besides, critical decision making for both humans and autonomous systems also demands a robust understanding and proactive forecasting of 3D geometry and object semantics.
To bridge this gap, we propose two straightforward yet effective training methods that enable the model to reason future by considering these environmental cues without relying on manual annotations.

% importance of 3D -> problem of existing methods -> our solution
\textbf{Reasoning 3D Geometry by Predicting the Future.}
Accurate 3D geometry encapsulates the spatial position and structural shape of surrounding objects, serving as a critical prerequisite for the safe navigation of autonomous driving models.
A prevalent approach to incorporating this 3D context is to supply the driving model with feature representations or outputs derived from pretrained depth estimation networks as additional inputs~\cite{vggdrive, world4drive}.
However, relying on these auxiliary models for depth-aware inputs introduces extra computational overhead, which inevitably increases the overall trajectory inference latency of the system.
Furthermore, without explicitly enforcing the utilization of these depth-aware inputs through task-specific constraints, it remains ambiguous whether the driving model genuinely internalizes these 3D features to infer the future for its planning decisions.
To address these concerns, we task our model with predicting the next frame metric depth $\hat{\bm{d}}_{i + 1}$ and a corresponding confidence map $\hat{\bm{c}}_{i + 1}$.
This is achieved by decoding the inferred future representations $F_i'$ to metric depth maps through a lightweight depth head $f_d$:
\begin{align}
    \{\hat{\bm{d}}_{i + 1}, \hat{\bm{c}}_{i + 1} \} = f_d(F_i', \Delta \hat{A}_{i+1}),
\end{align}
where $\Delta \hat{A}_{i + 1}$ represents the controlling relative movement predicted by our flow matching planner.
To supervise the depth decoder, we adopt a framework inspired by the visual foundation model Depth-Anything-V3~\cite{depth-anything-v1, depth-anything-v2, depth-anything-v3}:
\begin{align}\label{equa:loss-depth}
    L_d =\ & \hat{\bm{c}}_{i + 1}||\hat{\bm{d}}_{i + 1} - \bm{d}_{i + 1}||_1 - \lambda_c\log \hat{\bm{c}}_{i + 1} \notag \\ 
    &+ ||\nabla_x\hat{\bm{d}}_{i + 1} - \nabla_x\bm{d}_{i + 1}||_1 + ||\nabla_y\hat{\bm{d}}_{i + 1} - \nabla_y\bm{d}_{i + 1}||_1.
\end{align}
In this formulation, $\bm{d}_{i+1}$ serves as a pseudo metric depth label for the next frame, generated by Depth-Anything-V3~\cite{depth-anything-v3}.
Following Metric3D~\cite{metric3d, metric3dv2}, we scale the metric depth using a canonical focal length.
This loss allows our model to fully consider the 3D structures and object movements during future reasoning, ultimately enhancing planning performance without incurring extra computational costs during trajectory inference.

% importance of semantic -> SAM has semantic -> our solution
\textbf{Reasoning Semantic Information by Predicting the Future.}
Another critical factor for effective decision making is the semantic understanding of surrounding objects.
While pretrained visual foundation models for segmentation~\cite{sam3} can extract rich semantic information from visual inputs, previous works~\cite{world4drive} typically supervise the image backbone via an auxiliary current-frame segmentation task.
This approach captures limited semantic context and struggles to explicitly force the driving model to utilize this understanding for future inference and trajectory planning.
Therefore, to empower our model to autonomously internalize these semantics, we introduce a novel formulation that explicitly predicts the future-frame feature maps generated by SAM3~\cite{sam3}.
Given an object description, we obtain the encoded text features $\bm{h}_{text}$ from the SAM3 text branch and the corresponding text-fused image features $\bm{h}_i$.
Similar to our depth module, we utilize a lightweight head $f_{sam}$ to extract the text-aligned feature $\hat{\bm{h}}_{i + 1}$ for the next frame from the inferred future representation $F_i'$ with the text feature $\bm{h}_{text}$ from the text branch of SAM3:
\begin{align}\label{equa:sam-predict}
    \hat{\bm{h}}_{i + 1} = f_{sam}(F_i', \Delta \hat{A}_{i+1}, \bm{h}_{text}).
\end{align}
To facilitate semantic reasoning, we design the following loss function:
\begin{align}
    L_s = ||\hat{\bm{h}}_{i + 1} - \bm{h}_{i + 1}||^2.
\end{align}
In this equation, $\bm{h}_{i + 1}$ represents the pseudo next frame feature label generated by SAM3, fused by the text prompt $\bm{h}_{text}$ through the SAM3 text encoder.
Additionally, by formulating specific text descriptions for various objects in the driving environment, we can direct the attention of our model toward these critical elements.
This results in stronger real-world reasoning capabilities and richer future states information for trajectory planning.
To achieve faster convergence during training, we replace the predicted controlling movement $\Delta\hat{A}_{i + 1}$ in both the depth and semantic branches with the ground truth movement.
% Further technical details regarding our depth and semantic prediction modules are available in Section~\ref{sec:suppl-impl} of the supplementary material.

\subsection{Flow Matching Group Relative Policy Optimization for Planner}

Recent studies have demonstrated that reinforcement learning (RL) techniques are highly effective in enhancing the performance of large models, including planning accuracy~\cite{autovla, recogdrive}.
To further augment the planning capabilities of our model, we incorporate GRPO into our flow matching planner.
Following the Flow-GRPO framework~\cite{flow-grpo}, we reformulate the sampling process of the flow matching model as a Stochastic Differential Equation~(SDE):
\begin{align}\label{equa:flow-grpo}
    \hat{\bm{x}}_{t+\Delta t} = \underbrace{\hat{\bm{x}}_t (1 + \frac{\sigma_t^2}{2t}\Delta t) + v_{traj}(F_i', \Delta A_i', \hat{\bm{x}}_t, t)\cdot(1 + \frac{\sigma_t^2(1 - t)}{2t})\Delta t}_{\mu_{\theta}(\hat{\bm{x}}_{t}, t)} + \sigma_t\sqrt{|\Delta t|}\epsilon, \ \epsilon \sim \mathcal{N}(\mathbf{0}, \mathbf{I}),
\end{align}
where $\sigma_t = a\sqrt{\frac{t}{1-t}}$, and $a$ is a scalar hyper-parameter controlling the noise level.
In this formulation, the entire diffusion sampling process can be viewed as a continuous decision process~\cite{dppo}, where each diffusion step represents a Gaussian policy $\pi_{\theta}(\hat{\bm{x}}_{t+\Delta t} | \hat{\bm{x}}_{t})=\mathcal{N}(\mu_{\theta}(\hat{\bm{x}}_{t}, t), \sigma_t^2|\Delta t|\mathbf{I})$.
After generating a group of trajectory $\{\hat{\bm{x}}_0^g\}_{g=1}^G$ using Equation~\ref{equa:flow-grpo}, we can compute the group-standardized advantage for each trajectory:
\begin{align}\label{equa:group}
    A_g=\frac{r_g - mean(\{r_g\}_{g=1}^G)}{std(\{r_g\}_{g=1}^G)},
\end{align}
where $r_g$ is the reward of the trajectory $\hat{\bm{x}}_0^g$.
% !!!!!!!!!!!!!!!!!!!!!!! Remember to add reward function here !!!!!!!!!!!!!!!!!!!!!!!!!!!!!!!!!
Directly utilizing PDMS as the reward is unsuitable, as calculating PDMS requires manual perception labels~\cite{recogdrive, autovla}.
Therefore, we apply the generated trajectory to control a simulated vehicle, defining the reward based on the discrepancy between the ground-truth trajectory and the vehicle's simulated trajectory.
% !!!!!!!!!!!!!!!!!!!!!!!!!!!!!!!!!!!!!!!!!!!!!!!!!!!!!!!!!!!!!!!!!!!!!!!!!!!!!!!!!!!!!!!!!
The policy loss for the diffusion chain is then formulated as:
\begin{align}\label{equa:loss-rl}
    L_{rl} = -\frac{1}{G}\sum_{g=1}^G{\frac{1}{T}\sum_{t=1}^T{\gamma^{t-1}\log\pi_{\theta}(\hat{\bm{x}}_{(t-1)}^g|\hat{\bm{x}}_{(t)}^g)A_g}},
\end{align}
where $\gamma$ is the discount factor, $T$ is the total number of diffusion steps, and $\hat{\bm{x}}_{(t)}^g$ is the decision at each diffusion step for each group.
To stabilize the reinforcement learning process, we introduce an imitation learning loss to regularize the output of our flow matching planner:
\begin{align}\label{equa:loss-il}
    L_{il} = ||v_{traj}(F_i', \Delta A_i', \hat{\bm{x}}_t, t) - \frac{\hat{\bm{x}}_t -\bm{x}_0}{t}||^2,
\end{align}
where $\bm{x}_0$ is the ground truth trajectory.
This imitation loss constrains our planner to replicate the ground-truth trajectories from the dataset.
Overall, the final GRPO loss is defined as $L_{grpo}=L_{rl} + \lambda_{il}L_{il}$.
% The details of flow matching GRPO are in Section~\ref{sec:suppl-impl} of the supplementary material.

\textbf{Optimization.}
The optimization of \modelname~proceeds in two stages.
In the first stage, we focus on establishing the model's future reasoning ability of the real-world environment.
Specifically, we freeze the pretrained DINO-Tok and optimize the world model backbone, the depth head $f_d$, the semantic feature head $f_{sam}$, and the rectified flow matching heads $v_{traj}, v_{img}$.
The objective function for this first stage is formulated as $L=L_{traj} + L_{img} + L_d + L_s$.
In the second stage, we focus on fine-tuning the trajectory planner to achieve higher planning accuracy.
To this end, we freeze all components of the model except for the rectified flow matching planner $v_{traj}$, which is then fine-tuned using the GRPO loss $L_{grpo}$.
The pipeline of \modelname~is illustrated in Figure~\ref{fig:pipeline}.

\begin{table}[!t]
\caption{Comparison on the NAVSIMv1~\cite{navsim-v1} benchmark. `1xCam' means the front-view camera. `NxCam' means the surrounding-view cameras. `L' means Lidar.}
% $^*$ denotes the perception-free model is assisted by the trajectory anchor priors.}
\label{tab:navtest-v1}
\centering
\setlength{\tabcolsep}{1.2mm}{
\begin{tabular}{lcccccccc}
\toprule
Method & Perception & Input & NC & DAC & EP & TTC & C & PDMS \\
\midrule
Human & - & - & 100 & 100 & 87.5 & 100 & 99.9 & 94.8 \\
\midrule
% \color{gray}UniAD~\cite{uniad} & \color{gray}$\checkmark$ & \color{gray}6x Cam & \color{gray}97.8 & \color{gray}91.9 & \color{gray}78.8 & \color{gray}92.9 & \color{gray}100 & \color{gray}83.4 \\
\color{gray}Tranfuser~\cite{tranfusers-v1} & \color{gray}$\checkmark$ & \color{gray}3xCam + L & \color{gray}97.7 & \color{gray}92.8 & \color{gray}79.2 & \color{gray}92.8 & \color{gray}100 & \color{gray}84.0 \\
\color{gray}DiffusionDrive~\cite{diffusiondrive} & \color{gray}$\checkmark$ & \color{gray}3xCam + L & \color{gray}98.2 & \color{gray}96.2 & \color{gray}82.2 & \color{gray}94.7 & \color{gray}100 & \color{gray}88.1 \\
\color{gray}VGGDrive~\cite{vggdrive} & \color{gray}$\checkmark$ & \color{gray}3xCam & \color{gray}98.6 & \color{gray}96.3 & \color{gray}82.9 & \color{gray}95.6 & \color{gray}100 & \color{gray}88.8 \\
\color{gray}ResWorld~\cite{resworld} & \color{gray}$\checkmark$ & \color{gray}6xCam + L & \color{gray}98.9 & \color{gray}96.5 & \color{gray}83.1 & \color{gray}95.6 & \color{gray} 100 & \color{gray}89.0 \\
\color{gray}MeanFuser~\cite{meanfuser} & \color{gray}$\checkmark$ & \color{gray}3xCam & \color{gray}98.6 & \color{gray}97.0 & \color{gray}82.8 & \color{gray}95.0 & \color{gray}100 & \color{gray}89.0 \\
\color{gray}AutoVLA~\cite{autovla} & \color{gray}$\checkmark$ & \color{gray}3x Cam & \color{gray}98.4 & \color{gray}95.6 & \color{gray}81.9 & \color{gray}98.0 & \color{gray}99.9 & \color{gray}89.1 \\
\color{gray}DriveVLA-W0~(Anchor)~\cite{drivevla-w0} & \color{gray}$\checkmark$ & \color{gray}1xCam & \color{gray}98.7 & \color{gray}99.1 & \color{gray}83.3 & \color{gray}95.3 & \color{gray}99.3 & \color{gray}90.2 \\
\color{gray}AutoDrive-P$^3$~\cite{autodrive-p3} & \color{gray}$\checkmark$ & \color{gray}3xCam & \color{gray}99.1 & \color{gray}97.4 & \color{gray}84.8 & \color{gray}96.5 & \color{gray}100 & \color{gray} 90.6 \\
\color{gray}ReCogDrive~\cite{recogdrive} & \color{gray}$\checkmark$ & \color{gray}1xCam & \color{gray}97.9 & \color{gray}97.3 & \color{gray}87.3 & \color{gray}94.9 & \color{gray}100 & \color{gray}90.8 \\
\color{gray}SafeDrive~\cite{safedrive} & \color{gray}$\checkmark$ & \color{gray}3xCam + L & \color{gray}99.5 & \color{gray}99.0 & \color{gray}84.3 & \color{gray}97.2 & \color{gray}100 & \color{gray}91.6 \\
% \midrule
% \color{gray}World4Drive~\cite{world4drive}$^*$ & & \color{gray}3xCam & \color{gray}97.4 & \color{gray}94.3 & \color{gray}79.9 & \color{gray}92.8 & \color{gray}100 & \color{gray}85.1 \\
% \color{gray}DriveVLA-W0~(Anchor)~\cite{drivevla-w0}$^*$ & & \color{gray}1xCam & \color{gray}98.7 & \color{gray}99.1 & \color{gray}83.3 & \color{gray}95.3 & \color{gray}99.3 & \color{gray}90.2 \\
\midrule
DrivingGPT~\cite{drivinggpt} & & 1xCam & 98.9 & 90.7 & 79.7 & 94.9 & 95.6 & 82.4 \\
World4Drive~\cite{world4drive} & & 3xCam & 97.4 & 94.3 & 79.9 & 92.8 & \textbf{100} & 85.1 \\
Epona~\cite{epona} & & 1xCam & 97.9 & 95.1 & 80.4 & 93.8 & 99.9 & 86.2 \\
DriveVLA-W0~(Flow)~\cite{drivevla-w0} & & 1xCam & 98.4 & 95.3 & 80.9 & 95.2 & \textbf{100} & 87.2 \\
PWM~\cite{pwm} & & 1xCam & 98.6 & 95.9 & 81.8 & 95.4 & \textbf{100} & 88.1 \\
DriveVLA-W0~(Query)~\cite{drivevla-w0} & & 1xCam & 98.7 & 96.2 & 82.2 & 95.5 & \textbf{100} & 88.4 \\
DriveLaW~\cite{drivelaw} & & 1xCam & \textbf{99.0} & 97.1 & 81.3 & \textbf{96.7} & \textbf{100} & 89.1 \\
\modelname~(Ours) & & 1xCam & 98.6 & \textbf{97.9} & \textbf{84.8} & 95.7 & \textbf{100} & \textbf{90.4} \\
\bottomrule
\end{tabular}}
\end{table}

\begin{table}[!t]
\caption{Comparison on the NAVSIMv2~\cite{navsim-v2}~(navhard split) benchmark with human penalty \textbf{enabled}.}
\label{tab:navhard-v2}
\centering
\setlength{\tabcolsep}{0.7mm}{
\begin{tabular}{lcccccccccccc}
\toprule
Method & Perception & Stage & NC & DAC & DDC & TLC & EP & TTC & LK & HC & EC & EPDMS \\
\midrule
\color{gray}\multirow{2}{*}{LTF~\cite{transfuser}} & \color{gray}\multirow{2}{*}{$\checkmark$} & \color{gray}1 & \color{gray}96.2 & \color{gray}79.6 & \color{gray}99.1 & \color{gray}99.6 & \color{gray}84.1 & \color{gray}95.1 & \color{gray}94.2 & \color{gray}97.6 & \color{gray}79.1 & \color{gray}\multirow{2}{*}{25.1}\\
 & & \color{gray}2 & \color{gray}77.8 & \color{gray}70.2 & \color{gray}84.3 & \color{gray}98.1 & \color{gray}85.1 & \color{gray}75.7 & \color{gray}45.4 & \color{gray}95.7 & \color{gray}76.0 & \\
\midrule
\color{gray}\multirow{2}{*}{LTFv6~\cite{lead}} & \color{gray}\multirow{2}{*}{$\checkmark$} & \color{gray}1 & \color{gray}96.6 & \color{gray}86.7 & \color{gray}99.2 & \color{gray}99.6 & \color{gray}84.5 & \color{gray}95.1 & \color{gray}94.4 & \color{gray}97.8 & \color{gray}76.4 & \color{gray}\multirow{2}{*}{31.9} \\
 & & \color{gray}2 & \color{gray}79.9 & \color{gray}75.6 & \color{gray}86.3 & \color{gray}97.9 & \color{gray}89.6 & \color{gray}76.1 & \color{gray}50.1 & \color{gray}95.2 & \color{gray}66.7 & \\
\midrule
\color{gray}\multirow{2}{*}{RAP~\cite{rap}} & \color{gray}\multirow{2}{*}{$\checkmark$} & \color{gray}1 & \color{gray}97.1 & \color{gray}94.4 & \color{gray}98.8 & \color{gray}99.8 & \color{gray}83.9 & \color{gray}96.9 & \color{gray}94.7 & \color{gray}96.4 & \color{gray}66.2 & \color{gray}\multirow{2}{*}{39.6} \\
 & & \color{gray}2 & \color{gray}83.2 & \color{gray}83.9 & \color{gray}87.4 & \color{gray}98.0 & \color{gray}86.9 & \color{gray}80.4 & \color{gray}52.3 & \color{gray}95.2 & \color{gray}52.4 & \\
\midrule
\multirow{2}{*}{DriveVLA-W0~\cite{drivevla-w0}} & & 1 & 96.8 & 83.3 & 99.0 & 99.6 & \textbf{84.6} & 95.3 & 96.4 & 97.6 & \textbf{78.2} & \multirow{2}{*}{24.4} \\
 & & 2 & 76.8 & 64.3 & 79.9 & 98.3 & \textbf{89.2} & 75.0 & 46.8 & 95.8 & 53.1 & \\
\midrule
\multirow{2}{*}{DriveLaW~\cite{drivelaw}} & & 1 & \textbf{97.3} & 89.1 & 99.2 & 99.6 & 84.3 & 97.1 & 96.2 & \textbf{97.8} & 67.6 & \multirow{2}{*}{30.6} \\
 & & 2 & 82.5 & 67.6 & 83.5 & 98.1 & 84.8 & 78.5 & 45.8 & \textbf{96.4} & \textbf{57.3} & \\
\midrule
% \multirow{2}{*}{\modelname~(Ours)} & & 1 & 96.8 & 90.0 & 99.0 & 100 & 84.4 & 96.2 & 97.1 & 97.8 & 61.8 & \multirow{2}{*}{\textbf{35.0}} \\
%  & & 2 & 81.7 & 77.4 & 87.6 & 98.9 & 87.3 & 77.3 & 49.5 & 96.3 & 46.3 & \\
% \multirow{2}{*}{\modelname~(Ours)} & & 1 & 97.3 & \textbf{91.1} & \textbf{99.4} & \textbf{100} & 83.3 & \textbf{97.6} & \textbf{97.6} & 97.6 & 62.2 & \multirow{2}{*}{\textbf{36.4}} \\
%  & & 2 & \textbf{83.6} & \textbf{78.0} & \textbf{88.1} & \textbf{98.7} & 86.1 & \textbf{80.2} & \textbf{49.6} & \textbf{96.8} & 50.9 & \\
\multirow{2}{*}{\modelname~(Ours)} & & 1 & \textbf{97.3} & \textbf{90.7} & \textbf{99.4} & \textbf{100} & 83.3 & \textbf{97.3} & \textbf{97.3} & 97.6 & 60.9 & \multirow{2}{*}{\textbf{36.1}} \\
 & & 2 & \textbf{83.6} & \textbf{78.0} & \textbf{88.0} & \textbf{98.9} & 86.0 & \textbf{80.3} & \textbf{50.1} & 96.1 & 52.0 \\

\bottomrule
\end{tabular}}
\end{table}

\section{Experiments}\label{sec:experiments}

\subsection{Setup}

\textbf{Benchmarks.}
To evaluate our model, we select three challenging benchmarks from the NAVSIM~\cite{navsim-v1, navsim-v2} dataset: the `navtest' split evaluated under NAVSIMv1 metrics, the `navtest' split under NAVSIMv2 metrics, and the `navhard' split under NAVSIMv2 metrics.
The primary metric for NAVSIMv1~\cite{navsim-v1} is PDMS, which comprehensively assesses the model's performance across several dimensions: no at-fault collision~(NC), drivable area compliance~(DAC), ego progress~(EP), time-to-collision~(TTC) and comfort~(C).
The primary metric for NAVSIMv2~\cite{navsim-v2} is EPDMS.
In addition to the base PDMS components, EPDMS incorporates traffic light compliance~(TLC) and lane keeping~(LK), while expanding comfort~(C) into history comfort~(HC) and extended comfort~(EC).
For all reported metrics, higher values indicate better performance.

\textbf{Implementation.}
We adopt the language part of Qwen3-VL 4B~\cite{qwen3-vl} to initialize our world model backbone.
Totally, \modelname~has 6.7B parameters.
The input video resolution and frame rate are set to $512\times1024$ and 2Hz respectively.
The text features $\bm{h}_{text}$ in Equation~\ref{equa:sam-predict} are extracted via the SAM3~\cite{sam3} text encoder using `car' and `human' as prompts.
The entire training process requires approximately 10 days and is distributed across 64 H20 GPUs.
During the first stage, we train \modelname~on the nuPlan~\cite{nuplan} and nuScenes~\cite{nuscenes} datasets using 5-frame video clips to establish the model's foundational future reasoning ability of the real world.
In the second stage, we exclusively finetune the trajectory planner on the `navtrain' split of the NAVSIM dataset, utilizing navigational commands and 4-frame video inputs.
This stage incorporates our proposed flow matching GRPO to further enhance planning accuracy.
The evaluation of our model is conducted on a single RTX 4090 GPU.
% More details and experiments can be found in Section~\ref{sec:suppl-impl} and~\ref{sec:suppl-experiments} of the supplementary material.

\begin{table}[!t]
\caption{Comparison on the NAVSIMv2~\cite{navsim-v2}~(navtest split) benchmark with human penalty \textbf{enabled}.}
\label{tab:navtest-v2}
\centering
\setlength{\tabcolsep}{0.9mm}{
\begin{tabular}{lccccccccccc}
\toprule
Method & Perception & NC & DAC & DDC & TLC & EP & TTC & LK & HC & EC & EPDMS \\
\midrule
Human & - & 100 & 100 & 99.8 & 100 & 87.4 & 199 & 199 & 98.1 & 90.1 & 94.5 \\
\midrule
\color{gray}Tranfuser~\cite{tranfusers-v1} & \color{gray}$\checkmark$ & \color{gray}96.9 & \color{gray}89.9 & \color{gray}97.8 & \color{gray}99.7 & \color{gray}87.1 & \color{gray}95.4 & \color{gray}92.7 & \color{gray}98.3 & \color{gray}87.2 & \color{gray}84.0 \\
\color{gray}WoTE~\cite{wote} & \color{gray}$\checkmark$ & \color{gray}98.5 & \color{gray}96.8 & \color{gray}98.8 & \color{gray}99.8 & \color{gray}86.1 & \color{gray}97.9 & \color{gray}95.5 & \color{gray}98.3 & \color{gray}82.9 & \color{gray}87.7 \\
\color{gray}DiffusionDrive~\cite{diffusiondrive} & \color{gray}$\checkmark$ & \color{gray}98.2 & \color{gray}96.2 & \color{gray}99.5 & \color{gray}99.8 & \color{gray}87.4 & \color{gray}97.3 & \color{gray}96.9 & \color{gray}98.4 & \color{gray}87.7 & \color{gray}88.2 \\
\color{gray}AutoDrive-P$^3$~\cite{autodrive-p3} & \color{gray}$\checkmark$ & \color{gray}99.1 & \color{gray}97.4 & \color{gray}99.2 & \color{gray}99.8 & \color{gray}88.0 & \color{gray}98.7 & \color{gray}96.3 & \color{gray}98.3 & \color{gray}85.5 & \color{gray}89.9 \\
\midrule
DriveVLA-W0~\cite{drivevla-w0} & & 98.4 & 95.2 & 99.4 & \textbf{99.9} & 86.6 & 97.9 & \textbf{97.8} & 98.3 & 82.7 & 86.9 \\
PWM~\cite{pwm} & & \textbf{98.8} & 95.9 & 99.4 & \textbf{99.9} & 86.4 & \textbf{98.4} & 97.6 & 98.3 & \textbf{85.3} & 88.2 \\
DriveLaW~\cite{drivelaw} & & 98.7 & 96.9 & \textbf{99.6} & 99.8 & 87.5 & 98.3 & 97.6 & \textbf{98.4} & 77.4 & 88.6 \\
% \modelname~(Ours) & & 98.5 & \textbf{97.6} & \textbf{99.6} & \textbf{99.9} & \textbf{87.7} & 98.2 & \textbf{97.8} & 98.2 & 74.9 & \textbf{88.8} \\
\modelname~(Ours) & & 98.5 & \textbf{97.4} & 99.5 & \textbf{99.9} & \textbf{87.9} & 98.1 & 97.7 & 98.2 & 77.4 & \textbf{88.9} \\
\bottomrule
\end{tabular}}
\end{table}

\begin{table}[!t]
\caption{Ablation of the future prediction on the NAVSIMv1~\cite{navsim-v1} benchmark. The models are downscaled by using Qwen3-VL~\cite{qwen3-vl} 2B as the world model backbone.}
\label{tab:ablation-preception}
\centering
\setlength{\tabcolsep}{2.5mm}{
\begin{tabular}{cccccccccc}
\toprule
$L_{traj}$ & $L_{img}$ & $L_{d}$ & $L_{s}$ & NC & DAC & EP & TTC & C & PDMS \\
\midrule
$\checkmark$ & & & & 97.0 & 94.5 & 79.7 & 91.5 & 99.9 & 84.4 \\
$\checkmark$ & $\checkmark$ & & & 96.6 & 95.4 & 79.9 & 91.2 & 99.9 & 84.8 \\
$\checkmark$ & & $\checkmark$ & & 97.3 & 95.0 & 79.4 & 93.2 & 99.9 & 85.2 \\
$\checkmark$ & $\checkmark$ & $\checkmark$ & & 97.1 & 95.4 & 80.6 & 92.3 & 99.9 & 85.6 \\
$\checkmark$ & $\checkmark$ & $\checkmark$ & $\checkmark$ & \textbf{97.7} & \textbf{96.1} & \textbf{81.4} & \textbf{93.5} & \textbf{100} & \textbf{86.9} \\
\bottomrule
\end{tabular}}
\end{table}

\subsection{Comparison}

\textbf{Comparison on the NAVSIMv1~\cite{navsim-v1} Benchmark.}
Table~\ref{tab:navtest-v1} presents a comparison of our model with SOTA perception-based and perception-free models.
We classify DriveVLA-W0~(Anchor)~\cite{drivevla-w0} as a perception-based model because it applies manual perception labels to PDMS calculation for each trajectory anchor and uses the PDMS results to choose trajectories for supervision.
% Therefore, we do not include DriveVLA-W0~(Anchor)~\cite{drivevla-w0} as a fair perception-free baseline.
As shown in Table~\ref{tab:navtest-v1}, despite not relying on any manual perception labels, our model achieves planning performance comparable to that of perception-based models.
% Furthermore, owing to our proposed robust future self-learning mechanisms and GRPO-based refinements, our model preserves decision diversity across a wide range of driving scenarios.
Compared to other perception-free models, our model significantly outperforms Drive-VLA-W0~\cite{drivevla-w0} and DriveLaW~\cite{drivelaw}, achieving a notable \textit{\textbf{+1.3}} improvement in PDMS.
% Compared to other perception-free models, our model significantly outperforms the larger Drive-VLA-W0~\cite{drivevla-w0} and DriveLaW~\cite{drivelaw}, which contains over 9.2B and 8.0B parameters respectively.
These results demonstrate that incorporating future depth and semantic predictions, along with fine-tuning via flow matching GRPO, empowers driving models to better understand the environment and enables the planner to generate more accurate trajectories.

\textbf{Comparison on the NAVSIMv2~\cite{navsim-v2} Benchmark.}
The NAVSIMv2~\cite{navsim-v2} benchmark employs a two-stage pseudo closed-loop evaluation.
The corresponding results are presented in Table~\ref{tab:navtest-v2} and Table~\ref{tab:navhard-v2}.
For a fair comparison, we additionally evaluate DriveVLA-W0~(Flow)~\cite{drivevla-w0}, PWM~\cite{pwm} and DriveLaW~\cite{drivelaw} on these benchmarks using their publicly available checkpoints.
\modelname~achieves superior performance than other perception-free models on these more challenging benchmarks, which feature longer simulation horizons.
Notably, on the `navhard' split of NAVSIMv2 benchmark, our model has \textbf{\textit{+5.5}} improvement in EPDMS.
% \modelname~ with 6.7B parameters achieves superior performance than 8.0B DriveLaW on these more challenging benchmarks, which feature longer simulation horizons.
The higher EPDMS underscores our model's robust reasoning capabilities across diverse driving scenarios.

\subsection{Ablation Study}

\textbf{Ablation of Future Prediction Losses.}
The prediction losses task the model with forecasting future depth, images, and SAM3-generated semantic maps~\cite{sam3}, thereby establishing the ability of our model to comprehend the real-world environment and predict the future for trajectory planning.
Due to the high computational cost associated with training the full-scale model, we conduct an ablation study on these losses by downscaling \modelname~to 3.5B parameters using the Qwen3-VL~\cite{qwen3-vl} 2B as the world model backbone.
The results of progressively integrating these prediction losses are presented in Table~\ref{tab:ablation-preception}.
The results demonstrate that the depth loss enhances the model's understanding of 3D geometry, yielding notable improvements in NC, DAC and TTC.
Furthermore, the semantic map loss facilitates the extraction of semantic information from surrounding objects, which effectively prevents collisions and guides \modelname~to maintain safe, rule-compliant driving trajectories.

\begin{table}[!t]
\caption{Comparison of predicting the future and the present on the NAVSIMv1~\cite{navsim-v1} Benchmark. The models are downscaled by using Qwen3-VL~\cite{qwen3-vl} 2B as the world model backbone and only have the depth predictor for self-supervision.}
\label{tab:ablation-predict}
\centering
\setlength{\tabcolsep}{2.5mm}{
\begin{tabular}{lcccccc}
\toprule
Prediction & NC & DAC & EP & TTC & C & PDMS \\
\midrule
Present Frame & 96.3 & 94.5 & 78.8 & 90.8 & 99.8 & 83.6 \\
Future Frame & \textbf{97.3} & \textbf{95.0} & \textbf{79.4} & \textbf{93.2} & \textbf{99.9} & \textbf{85.2} \\
\bottomrule
\end{tabular}}
\end{table}

\textbf{Predict the Present \textit{v.s.} the Future.}
The future-forecasting forces \modelname~to utilize depth and semantic information for better trajectory planning.
To further demonstrate its effectiveness, we conduct an experiment comparing the impact of predicting current versus future depth maps.
Specifically, the setting `Present Frame' utilizes the current depth map to supervise the model, and the `Future Frame' setting utilizes the future depth map.
In both configurations, we rely solely on the depth predictor to establish the model's real-world reasoning capabilities and downscale our model for faster training.
The corresponding results are presented in Table~\ref{tab:ablation-predict}.
The superior performance achieved by forecasting future frames indicates that this approach fosters a substantially stronger and more comprehensive real-world future reasoning ability for trajectory planning within the model.

\begin{table}[!t]
\caption{Ablation of the flow matching GRPO on the NAVSIMv1~\cite{navsim-v1} Benchmark.}
\label{tab:ablation-rl}
\centering
\setlength{\tabcolsep}{2.5mm}{
\begin{tabular}{lcccccc}
\toprule
Model & NC & DAC & EP & TTC & C & PDMS \\
\midrule
\modelname~w/o $L_{grpo}$ & \textbf{98.6} & 97.3 & 83.6 & 95.3 & 99.9 & 89.4 \\
% \modelname~(Ours) & \textbf{98.8} & \textbf{97.8} & \textbf{83.9} & \textbf{96.4} & 99.4 & \textbf{90.3} \\
\modelname~(Ours) & \textbf{98.6} & \textbf{97.9} & \textbf{84.8} & \textbf{95.7} & \textbf{100} & \textbf{90.4} \\
\bottomrule
\end{tabular}}
\end{table}

\textbf{Ablation of Flow Matching GRPO.}
The flow matching GRPO further optimizes the trajectory planner using reinforcement learning techniques.
To evaluate its impact, we conduct an ablation study where the GRPO loss $L_{grpo}$ is removed, denoted as the `\modelname~w/o $L_{grpo}$' setting.
The results of this ablation study are presented in Table~\ref{tab:ablation-rl}.
These findings demonstrate that the flow matching GRPO significantly enhances planning accuracy, as evidenced by notable improvements in the EP metric.

% \subsection{Visualization}

% \begin{figure}[!t]
%     \centering
%     \includegraphics[width=0.95\linewidth, trim=0 340 160 90, clip]{figures/visualization.pdf}
%     \caption{Visualization of \textbf{future} prediction results. The trajectory in green means the ground truth planning, and the orange one means the results of our model.}
%     \label{fig:visualization}
% \end{figure}

% \textbf{Visualization.}
% To further show the real world future reasoning ability of our model, we visualize several future prediction results of \modelname~in Figure~\ref{fig:visualization}.
% % The semantic maps in Figure~\ref{fig:visualization} are generated with the prompt `car' and visualized via Principal Component Analysis~(PCA).
% The semantic maps in Figure~\ref{fig:visualization} are generated with the prompt `car' and visualized via PCA.
% The accurate future frame prediction indicates that our model effectively captures the motion dynamics of surrounding objects.
% Concurrently, the forecasted depth maps illustrate the model's robust 3D estimation capabilities, which are crucial for informed driving decisions.
% Furthermore, the semantic maps highlight the specific objects our model focuses on, thereby facilitating more rational and precise planning.
\section{Conclusion}\label{sec:conclusion}

In this paper, we propose \modelname, a novel paradigm of a perception-free driving world model.
To equip our model with comprehensive real-world future reasoning capabilities, we designed losses that task the model with predicting the depth and semantic maps of future frames.
By forecasting this 3D and semantic information, \modelname~can effectively perceive the geometry of surrounding objects and comprehend their semantic context, resulting in enhanced trajectory planning performance.
Additionally, we introduce a flow matching group relative policy optimization mechanism to further refine our trajectory planner and improve overall planning accuracy.
Extensive experiments demonstrate the effectiveness of our proposed approach.
However, our model does not yet surpass SOTA perception-based models, primarily due to challenges arising from the inherent imprecision of the pseudo depth and semantic labels generated by pretrained foundational models.
We leave the resolution of these limitations to future work and we hope that \modelname~will inspire further exploration into perception-free models toward the realization of true general driving intelligence.

% \begin{ack}
% Use unnumbered first level headings for the acknowledgments. All acknowledgments

% Do {\bf not} include this section in the anonymized submission, only in the final paper. You can use the \texttt{ack} environment provided in the style file to automatically hide this section in the anonymized submission.
% \end{ack}

\clearpage
{
    \small
    \bibliographystyle{plain}
    \bibliography{references}
}

%%%%%%%%%%%%%%%%%%%%%%%%%%%%%%%%%%%%%%%%%%%%%%%%%%%%%%%%%%%%

% \clearpage
% \appendix
% \input{sections/X_supplementary}

% \clearpage
% \input{checklist.tex}

\end{document}